\documentclass{article}
\pdfoutput=1
\usepackage{blindtext}
\usepackage{graphicx}
\usepackage{amsmath}
\usepackage{amssymb}
\usepackage{array}
\usepackage{subfig}
\usepackage{hyperref}
\usepackage{shortcuts}
\usepackage{bm}
\usepackage[ruled,norelsize]{algorithm2e}

\def\adj{^{\top}}

\newtheorem{lemma}{Lemma}

\makeatletter
\newcommand*{\inlineequation}[2][]{%
  \begingroup
    \refstepcounter{equation}%
    \ifx\\#1\\%
    \else
      \label{#1}%
    \fi
    \relpenalty=10000 %
    \binoppenalty=10000 %
    \ensuremath{%
      #2%
    }%
    ~\@eqnnum
  \endgroup
}
\makeatother

\begin{document}

\title{Faster ICA under orthogonal constraint}

\author{Pierre Ablin \thanks{P. Ablin works at Inria, Parietal Team, Universit\'e Paris-Saclay, Saclay, France; e-mail: pierre.ablin@inria.fr},
        Jean-Francois Cardoso \thanks{J. F. Cardoso is with the Institut d'Astrophysique de Paris, CNRS, Paris, France; e-mail: cardoso@iap.fr},
        and Alexandre Gramfort 
\thanks{A. Gramfort is with Inria, Parietal Team, Universit\'e Paris-Saclay, Saclay, France; e-mail: alexandre.gramfort@inria.fr}}

\maketitle

\begin{abstract}

Independent Component Analysis (ICA) is a technique for unsupervised exploration of multi-channel data widely used in observational sciences. In its classical form, ICA relies on modeling the data as a linear mixture of non-Gaussian independent sources. The problem can be seen as a likelihood maximization problem. We introduce Picard-O, a preconditioned L-BFGS strategy over the set of orthogonal matrices, which can quickly separate both super- and sub-Gaussian signals. It returns the same set of sources as the widely used FastICA algorithm. Through numerical experiments, we show that our method is faster and more robust than FastICA on real data.

\end{abstract}


\section{Introduction}

Independent component analysis (ICA) \cite{comon1992independent} is a popular technique for multi-sensor signal processing. 
Given an $N\times T$ data matrix $X$ made of $N$ signals of length $T$, an ICA algorithm finds an $N\times N$ `unmixing matrix' $W$ such that the rows of $Y=WX$ are `as independent as possible'. An important class of ICA methods further constrains the rows of $Y$ to be uncorrelated.  Assuming zero-mean and unit variance signals, that is:
\begin{equation}
\label{eq:white}
\textstyle \frac 1T Y Y^\top = I_N \enspace .
\end{equation}
The `whiteness constraint'~\eqref{eq:white} is satisfied if, for instance,
\begin{equation}
\label{eq:Wpolar}
W = O W_0 \qquad W_0 = (\textstyle \frac 1T X X^\top)^{-1/2}
\end{equation}
and the matrix $O$ is constrained to be orthogonal: $ OO\adj=I_N$.
For this reason, ICA algorithms which enforce signal decorrelation by proceeding as suggested by \eqref{eq:Wpolar}  ---whitening followed by rotation--- can be called `orthogonal methods'.

The orthogonal approach is followed by many ICA algorithms.  Among them, 
FastICA~\cite{hyvarinen1999fast} stands out by its simplicity, good scaling and a built-in ability to deal with both sub-Gaussian and super-Gaussian components.  It also enjoys an impressive convergence speed when applied to data which are actual mixture of independent components ~\cite{oja2006fastica, shen2008local}.
However, real data can rarely be accurately modeled as mixtures of independent components. In that case, the convergence of FastICA may be impaired or even not happen at all.

In~\cite{ablin2017faster}, we introduced the Picard algorithm for fast non-orthogonal ICA on real data.   In this paper, we extend it to an orthogonal version, dubbed Picard-O, which solves the same problem as FastICA, but faster on real data.

In section~\ref{sec:problem}, the non-Gaussian likelihood for ICA is studied in the orthogonal case yielding the Picard Orthogonal (Picard-O) algorithm in section~\ref{sec:algo}.  Section~\ref{sec:fast} connects our approach to FastICA: Picard-O converges toward the fixed points of FastICA, yet faster thanks to a better approximation of the Hessian matrix.  
This is illustrated through extensive experiments on four types of data in section~\ref{sec:expe}.


\section{Likelihood under whiteness constraint}
\label{sec:problem}

Our approach is based on the classical non-Gaussian ICA likelihood~\cite{pham1997blind}. 
The $N\times T$ data matrix $X$ is modeled as $X=AS$ where the $N\times N$ mixing matrix $A$ is invertible and where $S$ has statistically independent rows: the `source signals'. Further, each row $i$ is modeled as an i.i.d. signal with $p_i(\cdot)$ the probability density function (pdf) common to all samples. In the following, this assumption is denoted as the mixture model. It never perfectly holds on real problems.
Under this assumption, the likelihood of $A$ reads:
\begin{equation}
\label{eq:loglika}
\textstyle
    p(X \lvert A) 
    = \prod_{t = 1}^{T} \frac{1}{\lvert \det(A) \rvert} 
    \prod_{i = 1}^{N}p_i([A^{-1} x]_i(t)) \enspace .
\end{equation}
It is convenient to work with the negative averaged log-likelihood parametrized by the unmixing matrix $W = A^{-1}$, that is, $\mathcal{L}(W) = -\frac1T \log p(X \lvert W^{-1})$. Then, \eqref{eq:loglika} becomes:
\begin{equation}
\textstyle
    \mathcal{L}(W) = -\log\lvert\det(W)\rvert -
    \hat{E}\left[\sum_{i = 1}^{N} \log(p_i(y_i(t))\right] \enspace ,
\label{eq:loglik}
\end{equation}
where $\hat{E}$ denotes the empirical mean (sample average) and where, implicitly, $[y_1,\cdots, y_N]^{\top} = Y=WX$.

\bigskip

We consider maximum likelihood estimation under the whiteness constraint~\eqref{eq:white}.
By \eqref{eq:Wpolar} and \eqref{eq:loglik}, this is equivalent to minimizing $\mathcal{L}(OW_0)$ with respect to the orthogonal matrix $O$.
To do so, we propose an iterative algorithm. A given iterate $W_k=O_k W_0$ is updated by replacing $O_k$ by a more likely orthonormal matrix $O_{k+1}$ in its neighborhood.
Following classical results of differential geometry over the orthogonal group~\cite{bertram:2008}, we parameterize that neighborhood by expressing $O_{k+1}$ as 
$O_{k+1} =e^{\mathcal{E}} O_k$ where $\mathcal{E}$ is a (small) $N\times N$ skew-symmetric matrix: $\mathcal{E}^\top = - \mathcal{E}$.

The second-order Taylor expansion of $\mathcal{L}(e^{\mathcal{E}}W)$ reads:
\begin{equation}
\mathcal{L}(e^{\mathcal{E}}W) = \mathcal{L}(W) + \langle G \lvert \mathcal{E}\rangle + \frac{1}{2} \langle \mathcal{E} \lvert H \rvert  \mathcal{E} \rangle +  \mathcal{O}(\lvert \lvert \mathcal{E} \rvert \rvert ^3).
\label{eq:taylor}
\end{equation}
The first order term is controlled by the $N\times N$ matrix $G=G(Y)$, called relative gradient and the second-order term depends on the $N\times N\times N \times N$ tensor $H(Y)$, the relative Hessian.
Both quantities only depend on $Y=WX$
and have simple expressions in the case of a small relative perturbation of the form $W \leftarrow (I + \mathcal{E}) W$ (see~\cite{ablin2017faster} for instance).  Those expressions are readily adapted to the case of interest,
using the second order expansion $\exp(\mathcal{E}) \approx I + \mathcal{E} +\frac12 \mathcal{E}^2$.  One gets:
\begin{align}
\label{eq:relatgrad}
G_{i j} 
&= \hat{E}[\psi_i(y_i)y_j] - \delta_{ij} \enspace , 
\\
H_{i j k l} 
&= \delta_{i l} \delta_{j k} \hat{E}[\psi_i(y_i)y_i]  + \delta_{ik} \, \hat{E}[\psi_i'(y_i)y_jy_l]  
\enspace .
\label{eq:hessian}
\end{align}
where $\psi_i = -\frac{p_i'}{p_i}$ is called the \emph{score function}.

This Hessian~\eqref{eq:hessian} is quite costly to compute, but a simple approximation is obtained using the form that it would take for large $T$ and independent signals.  %
In that limit, one has:
\begin{equation}
\hat{E}[\psi_i'(y_i)y_jy_l] 
\approx
\delta_{jl}\, \hat{E}[\psi_i'(y_i)]\hat{E}[y_j^2] 
\quad\text{for}\quad i \neq j,
\end{equation}
hence the approximate Hessian (recall $\hat{E}[y_j^2]=1$):
\begin{equation}
\tilde{H}_{i j k l} = \delta_{i l} \delta_{j k} \hat{E}[\psi_i(y_i)y_i] + \delta_{ik} \delta_{jl}\, \hat{E}[\psi_i'(y_i)]  \enspace \text{if } i \neq j.
\label{eq:approx}
\end{equation}
Plugging this expression in the 2nd-order expansion~\eqref{eq:taylor} and expressing the result as a function of the $N(N-1)/2$ free parameters $\{\mathcal{E}_{ij},\ 1\leq i< j\leq N\}$ of a skew-symmetric matrix $\mathcal{E}$ yields after simple calculations:
\begin{equation}
\langle G \lvert \mathcal{E}\rangle + \frac{1}{2} \langle \mathcal{E} \lvert \tilde H \rvert  \mathcal{E} \rangle 
=
\sum_{i< j}
(G_{ij}-G_{ji})\,\mathcal{E}_{ij}
+
\frac{\hat\kappa_i+\hat\kappa_j}{2}\, \mathcal{E}_{ij}^2
\label{eq:taylorskew}
\end{equation}
where we have defined the non-linear moments:
\begin{equation}
\hat\kappa_i 
= 
\hat{E}[\psi_i(y_i) y_i] - \hat{E}[\psi_i'(y_i)] \enspace .
\label{eq:kappa}
\end{equation}
If $\hat\kappa_i +\hat\kappa_j >0$ (this assumption will be enforced in the next section), the form~\eqref{eq:taylorskew} is minimized for $\mathcal{E}_{ij} = -(G_{ij}-G_{ji})/(\hat\kappa_i + \hat\kappa_j)$: 
the resulting quasi-Newton step would  be
\begin{equation}
W_{k+1}=e^{D}W_k
\quad\text{for}\quad
D_{ij} = -\frac{2}{\hat\kappa_i +\hat\kappa_j} \frac{G_{ij}-G_{ji}}{2}
.
\label{eq:Qnewt}
\end{equation}
That observation forms the keystone of the new orthogonal algorithm described in the next section.

\section{The Picard-O algorithm}
\label{sec:algo}

As we shall see in Sec.~\ref{sec:fast}, update~\eqref{eq:Qnewt} is essentially the behavior of FastICA near convergence.  
Hence, one can improve on FastICA by using a more accurate Hessian approximation.  Using the exact form~\eqref{eq:hessian}  would be quite costly for large data sets.
Instead, following the same strategy as~\cite{ablin2017faster}, we base our algorithm on the L-BFGS method  (which learns the actual curvature of the problem from the data themselves) using approximation~\eqref{eq:taylorskew} only as a \emph{pre-conditioner}.

\medskip
\textit{L-BFGS and its pre-conditioning:} The L-BFGS method keeps track of the $m$ previous values of the (skew-symmetric) relative moves $\mathcal{E}_k$ and gradient differences $\Delta_k = (G_k-G_k^\top)/2  - (G_{k-1}-G_{k-1}^\top)/2$ and of auxiliary quantities $\rho_k = \langle \mathcal{E}_k \lvert \Delta_k \rangle$.  It returns a descent direction by running through one backward loop and one forward loop.  It can be pre-conditioned by inserting a Hessian approximation in between the two loops as summarized in algorithm~\ref{algo:twoloop}.
\medskip
\textit{Stability:} If the ICA mixture model holds, the sources should constitute a local minimum of $\mathcal{L}$. According to~\eqref{eq:taylorskew}, that happens if $\hat\kappa_i + \hat\kappa_j > 0$ for all $i <j$ (see also~\cite{cardoso:stability:NNSP98}). 
We enforce that property by taking, at each iteration:
\begin{equation}
\psi_i(\cdot) = \sign(k_i) \psi(\cdot)
\label{eq:adaptpsi}
\end{equation}
where $k_i = \hat{E}[\psi(y_i)y_i] - \hat{E}[\psi'(y_i)]$ and $\psi$ is a fixed non-linearity (a typical choice is $\psi(u) = \tanh(u)$). 
This is very similar to the technique in extended Infomax~\cite{lee1999independent}.
It enforces $\hat\kappa_i = \lvert k_i \rvert >0$, and the positivity of $\tilde{H}$.  
In practice, if for any signal $i$ the sign of $k_i$ changes from one iteration to the next, L-BFGS's memory is flushed.

\medskip
\textit{Regularization:} The switching technique guarantees that the Hessian approximation is positive, but one may be wary of very small values of $\hat\kappa_i+\hat\kappa_j$.
Hence, the pre-conditioner uses a floor: $\max((\hat\kappa_i + \hat\kappa_j)/2, \kappa_\text{min})$ for some small positive value of $\kappa_{\text{min}}$ (typically $\kappa_\text{min}\simeq 10^{-2}$).

\medskip
\textit{Line search:} A backtracking line search helps convergence. Using the search direction $D_k$ returned by L-BFGS, and starting from a step size $\alpha = 1$, if  $\mathcal{L}(\exp(\alpha D_k) W_k) < \mathcal{L}(W_k)$, then set $\mathcal{E}_{k+1} = \alpha D_k$ and $W_{k+1} = \exp(\mathcal{E}_{k+1}) W_k$, otherwise divide $\alpha$ by 2 and repeat.

\medskip

\textit{Stopping:} The stopping criterion is $\lvert \lvert G -G^\top \rvert \rvert < \varepsilon$ where $\varepsilon$ is a small tolerance constant. 

\medskip

Combining these ideas, we obtain the Preconditioned Independent Component Analysis for Real Data-Orthogonal (Picard-O) algorithm,
summarized in table~\ref{algo:main}.

The Python code for Picard-O is available online at \url{https://github.com/pierreablin/picard}.

\begin{algorithm}[tb]
\SetKwInOut{Input}{Input}
\SetKwInOut{Output}{Output}
 \Input{Current gradient $G_k$, moments $\hat\kappa_i$, previous $\mathcal{E}_l$, $\Delta_l$, $\rho_l$ $\forall l \in \{k-m,\dots,k-1\}$.}
Set $Q = -(G_k-G_k^\top)/2$\;
 \For{l=k-1,\dots,k-m}{
	Compute $a_l = \rho_l \langle \mathcal{E}_l \lvert Q\rangle$ \;
    Set $Q = Q - a_l \Delta_i$ \;
  }
  Compute $D$ as $D_{ij} = Q_{ij} / \max\left(\frac{\hat\kappa_i + \hat\kappa_j}{2}, \kappa_\text{min}\right)$ \;
  \For{l=k-m,\dots,k-1}{
  	Compute $\beta = \rho_l \langle \Delta_l \lvert D \rangle$ \;
    Set $D = D + \mathcal{E}_l(a_l - \beta)$ \;
  }
 \Output{Descent direction $D$}
 \caption{Two-loop recursion L-BFGS formula}
 \label{algo:twoloop}
\end{algorithm}

\begin{algorithm}
\SetKwInOut{Input}{Input}
\SetKwInOut{Output}{Output}
 \Input{Initial signals $X$, number of iterations K}
 Sphering: compute $W_0$ by \eqref{eq:white} and set $Y=W_0 X$\;
 \For{$k = 0\cdots K$}{
  Compute the signs $\sign(k_i)$ \;
  Flush the memory if the sign of any source has changed \;
  Compute the gradient $G_k$ \;
  Compute search direction $D_k$ using algorithm~\ref{algo:twoloop} \;
  Compute the step size $\alpha_k$ by line search \;
  Set $W_{k+1} = \exp(\alpha_k D_k) W_k$ and $Y=W_{k+1} X$ \;
  Update the memory;
}
 \Output{Unmixed signals $Y$, unmixing matrix $W_k$}
 \caption{The Picard-O algorithm}
 \label{algo:main}
\end{algorithm}

\section{Link with FastICA}
\label{sec:fast}

This section briefly examines the connections between Picard-O and symmetric FastICA~\cite{hyvarinen1999fixed}.
In particular, we show that both methods essentially
share the same solutions and that the behavior of FastICA is similar to a quasi-Newton method. 


Recall that FastICA is based on an $N\times N$  matrix $C(Y)$ matrix defined entry-wise by:
\begin{equation}
 \label{eq:defC}
  C_{ij}(Y) = \hat{E}[\psi_i(y_i)y_j] - \delta_{ij} \hat{E}[\psi_i'(y_i)]  \enspace .
\end{equation}
The symmetric FastICA algorithm, starting from white signals $Y$,
can be seen as iterating $Y \leftarrow C_w(Y) Y$ until convergence,
where $C_w(Y)$ is the orthogonal matrix computed as
\begin{equation}
C_w(Y) = (C C^{\top})^{-\frac12} C \enspace .
\label{eq:sym}
\end{equation}
In the case of a fixed score function, a sign-flipping phenomenon appears leading to the following definition of a fixed point: $Y$ is a fixed point of FastICA if $C_w(Y)$ is a diagonal matrix of $\pm1$~\cite{wei2015convergence}.
This behavior can be fixed by changing the score functions as in~\eqref{eq:adaptpsi}.
It is not hard to see that, if $\psi$ is an odd function, such a modified version has the same trajectories as the fixed score version (up to to irrelevant sign flips), and that the fixed points of the original algorithm now all verify $C_w(Y) = I_N$.

\medskip

\textit{Stationary points : }
We first relate the fixed points of FastICA (or rather the sign-adjusted version described above) to the stationary points of Picard-O. 

Denote $C_+$ (resp. $C_-$) the symmetric (resp. skew-symmetric) part of $C(Y)$
and similarly for $G$.  It follows from~\eqref{eq:relatgrad} and~\eqref{eq:defC} that 
\begin{displaymath}
C = C_+ + C_- = C_+ + G_-
\end{displaymath}
since $C_-=G_- = (G - G^{\top})/2$.

One can show that $Y$ is a fixed point of FastICA if and only if $G(Y)$ is symmetric and $C_+(Y)$ is positive definite.
Indeed, at a fixed point, $C_w(Y)=I_N$, so that by Eq.~(\ref{eq:sym}), one has $C(Y) = (CC^\top)^{1/2}$ which is a positive matrix (almost surely).
Conversely, if $G(Y)$ is symmetric, then so is $C(Y)$.  If $C(Y)$ is also positive, then its polar factor $C_w(Y)$ is the identity matrix, so that $Y$ is a fixed point of FastICA.   

The modification of FastICA ensures that the diagonal of $C(Y)$ is positive, but does not guarantee positive definiteness.  However, we empirically observed that on each dataset used in the experiments, the matrix $C_+(Y)$ is positive definite when $G_-(Y)$ is small. 
Under that condition, we see that the stationary points of Picard-O, characterized by $G_-(Y)=0$ are exactly the fixed points of FastICA.

\medskip

\textit{Asymptotic behavior of FastICA : }
Let us now expose the behavior of FastICA close to a fixed point \textit{i.e.}  %
when $C_w(Y) = \exp({\mathcal{E}})$ for some small skew-symmetric matrix ${\mathcal{E}}$.

At first order in $\mathcal{E}$, the polar factor $C_w=\exp(\mathcal{E})$ of $C$ is obtained as solution of (proof omitted):
\begin{equation}
G_- = \frac{C_+\mathcal{E}+\mathcal{E}C_+}{2} \enspace .
\end{equation}
Denote by $\hat{H}$ the linear mapping $\hat{H} : \mathcal{E} \rightarrow -\frac{C_+\mathcal{E}+\mathcal{E}C_+}{2}$. 
When FastICA perform a small move, it is (at first order) of the form $W \leftarrow e^D W$ with $D = -\hat{H}^{-1}(G_-)$. It corresponds to a quasi-Newton step with $\hat{H}$ as approximate Hessian.

Furthermore, under the mixture model assumption, close from separation and with a large number of samples, $C_+$ becomes the diagonal matrix of coefficients $\delta_{ij}\hat\kappa_i$ and $\hat{H}$ simplifies, giving the same direction $D$ given in~\eqref{eq:Qnewt}.

In summary, we have shown that a slightly modified version of FastICA (with essentially the same iterates as the original algorithm) has the same fixed points as Picard-O. Close to such a fixed point, each of FastICA's iteration is similar to a quasi-Newton step with an approximate Hessian. This approximation matches the true Hessian if the mixture model holds, but this cannot be expected in practice on real data.

\section{Experiments}
\label{sec:expe}

This section illustrates the relative speeds of FastICA and Picard-O.
Both algorithms are coded in the Python programming language. Their computation time being dominated by score evaluations, dot products and sample averages, a fair comparison is obtained by ensuring that both algorithms call the exact same procedures so that speeds differences mostly stems from algorithmics rather than implementation.  

\begin{figure}
  \centering
  \includegraphics[width=1\columnwidth]{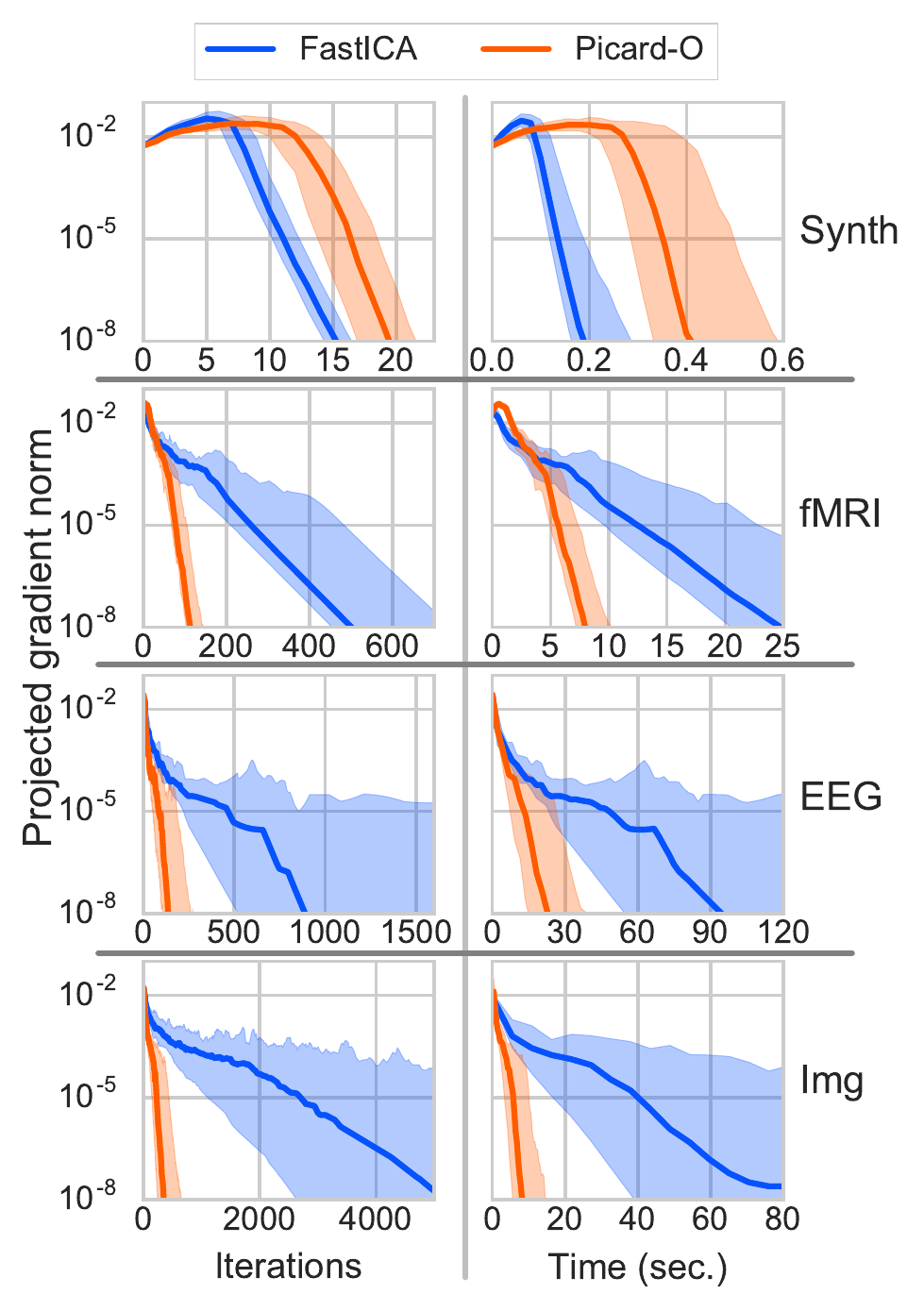}\hfill
  \caption{Comparison between FastICA and Picard-O. Gradient norm vs iterations (left column) and vs time (right column). From top to bottom: simulated data, fMRI data, EEG data and image data. Solid line corresponds to the median of the runs, the shaded area covers the $10\% - 90 \%$ percentiles.}
  \label{fig:expe}
\end{figure}

Figure~\ref{fig:expe} summarizes our results.
It shows the evolution of the projected gradient norm $\lvert \lvert G - G^{\top} \rvert \rvert$ versus iterations (left column) and versus time (right column). The 4 rows correspond to 4 data types: synthetic mixtures, fMRI signals, EEG signals, and image patches.
FastICA and Picard-O speeds are compared using $\psi(\cdot) = \tanh(\cdot)$.
The signals are centered and whitened before running ICA.

Experiments are repeated several times for each setup. The solid line shows the median of the gradient curves and the shaded area shows the $10~\%-90~\%$ percentile (meaning that half the runs completed faster than the solid line and that $80 \%$ have convergence curves in the shaded area).

\paragraph*{Synthetic data}
We generate $N = 50$ i.i.d. sources of length $T=10000$. The $25$ first sources follow a uniform law between $-1$ and $1$, the $25$ last follow a Laplace law ($p \propto \exp(-\lvert x \rvert)$). 
The $N\times T$ source matrix $S$ is multiplied by a random square mixing matrix $A$.
This experiment is repeated $100$ times, changing each time the seed generating the signals and the mixing matrix.

\paragraph*{fMRI}
This is functional MRI data processed by group ICA~\cite{varoquaux2010group}. The datasets come from ADHD-200 consortium~\cite{adhd2012adhd}. The problem is of size $N = 60$, $T=60000$, and the experiments are repeated over $20$ datasets.

\paragraph*{EEG}
ICA is applied on $13$ publicly available\footnote{https://sccn.ucsd.edu/wiki/BSSComparison} electroencephalography datasets~\cite{delorme2012independent}. Each recording contains $N=71$ signals, of length $T \simeq 75000$.

\paragraph*{Image patches}
We use a database of $80$ different images of open country~\cite{oliva2001modeling}.
From each image, $T=10000$ patches of size $8 \times 8$ are extracted and vectorized to obtain $N = 64$ signals, before applying ICA.

\medskip
\par\noindent\textbf{Results.}
FastICA is slightly faster than Picard-O on the simulated problem, for which the ICA mixture model holds perfectly.  However, on real data, the rate of convergence of FastICA is severely impaired because the underlying Hessian approximation is far from the truth, while our algorithm still converges quickly. Picard-O is also more consistent in its convergence pattern, showing less spread than FastICA. 



%



\section{Discussion}

In this paper, we show that, close from its fixed points, FastICA's iterations are similar to quasi-Newton steps for maximizing a likelihood. Furthermore, the underlying Hessian approximation matches the true Hessian of the problem if the signals are independent. However, on real datasets, the independence assumption never perfectly holds. Consequently, FastICA may converge very slowly on applied problems~\cite{chevalier2004comparative} or can get stuck in saddle points~\cite{tichavsky2006performance}.

To overcome this issue, we propose the Picard-O algorithm. As an extension of~\cite{ablin2017faster}, it uses a preconditioned L-BFGS technique to solve the same minimization problem as FastICA. Extensive experiments on three types of real data demonstrate that Picard-O can be orders of magnitude faster than FastICA.

\newpage

\bibliographystyle{IEEEtran}
\bibliography{bibliography}

\begin{thebibliography}{10}
\providecommand{\url}[1]{#1}
\csname url@samestyle\endcsname
\providecommand{\newblock}{\relax}
\providecommand{\bibinfo}[2]{#2}
\providecommand{\BIBentrySTDinterwordspacing}{\spaceskip=0pt\relax}
\providecommand{\BIBentryALTinterwordstretchfactor}{4}
\providecommand{\BIBentryALTinterwordspacing}{\spaceskip=\fontdimen2\font plus
\BIBentryALTinterwordstretchfactor\fontdimen3\font minus
  \fontdimen4\font\relax}
\providecommand{\BIBforeignlanguage}[2]{{%
\expandafter\ifx\csname l@#1\endcsname\relax
\typeout{** WARNING: IEEEtran.bst: No hyphenation pattern has been}%
\typeout{** loaded for the language `#1'. Using the pattern for}%
\typeout{** the default language instead.}%
\else
\language=\csname l@#1\endcsname
\fi
#2}}
\providecommand{\BIBdecl}{\relax}
\BIBdecl

\bibitem{comon1992independent}
P.~Comon, ``Independent component analysis, a new concept?'' \emph{Signal
  Processing}, vol.~36, no.~3, pp. 287 -- 314, 1994.

\bibitem{hyvarinen1999fast}
A.~Hyvarinen, ``Fast and robust fixed-point algorithms for independent
  component analysis,'' \emph{IEEE Transactions on Neural Networks}, vol.~10,
  no.~3, pp. 626--634, 1999.

\bibitem{oja2006fastica}
E.~Oja and Z.~Yuan, ``The fastica algorithm revisited: Convergence analysis,''
  \emph{IEEE Transactions on Neural Networks}, vol.~17, no.~6, pp. 1370--1381,
  2006.

\bibitem{shen2008local}
H.~Shen, M.~Kleinsteuber, and K.~Huper, ``Local convergence analysis of
  {FastICA} and related algorithms,'' \emph{IEEE Transactions on Neural
  Networks}, vol.~19, no.~6, pp. 1022--1032, 2008.

\bibitem{ablin2017faster}
P.~Ablin, J.-F. Cardoso, and A.~Gramfort, ``Faster independent component
  analysis by preconditioning with hessian approximations,'' \emph{Arxiv
  Preprint}, 2017.

\bibitem{pham1997blind}
D.~T. Pham and P.~Garat, ``Blind separation of mixture of independent sources
  through a quasi-maximum likelihood approach,'' \emph{IEEE Transactions on
  Signal Processing}, vol.~45, no.~7, pp. 1712--1725, 1997.

\bibitem{bertram:2008}
W.~Bertram, ``{Differential Geometry, Lie Groups and Symmetric Spaces over
  General Base Fields and Rings},'' \emph{{Memoirs of the American Mathematical
  Society}}, 2008.

\bibitem{cardoso:stability:NNSP98}
J.-F. Cardoso, ``On the stability of some source separation algorithms,'' in
  \emph{Proc. of the 1998 IEEE SP workshop on neural networks for signal
  processing (NNSP~'98)}, 1998, pp. 13--22.

\bibitem{lee1999independent}
T.-W. Lee, M.~Girolami, and T.~J. Sejnowski, ``Independent component analysis
  using an extended infomax algorithm for mixed subgaussian and supergaussian
  sources,'' \emph{Neural computation}, vol.~11, no.~2, pp. 417--441, 1999.

\bibitem{hyvarinen1999fixed}
A.~Hyv{\"a}rinen, ``The fixed-point algorithm and maximum likelihood estimation
  for independent component analysis,'' \emph{Neural Processing Letters},
  vol.~10, no.~1, pp. 1--5, 1999.

\bibitem{wei2015convergence}
T.~Wei, ``A convergence and asymptotic analysis of the generalized symmetric
  {FastICA} algorithm,'' \emph{IEEE transactions on signal processing},
  vol.~63, no.~24, pp. 6445--6458, 2015.

\bibitem{varoquaux2010group}
G.~Varoquaux, S.~Sadaghiani, P.~Pinel, A.~Kleinschmidt, J.-B. Poline, and
  B.~Thirion, ``A group model for stable multi-subject {ICA} on {fMRI}
  datasets,'' \emph{Neuroimage}, vol.~51, no.~1, pp. 288--299, 2010.

\bibitem{adhd2012adhd}
A.-. Consortium, ``The {ADHD-200} consortium: a model to advance the
  translational potential of neuroimaging in clinical neuroscience,''
  \emph{Frontiers in systems neuroscience}, vol.~6, 2012.

\bibitem{delorme2012independent}
A.~Delorme, J.~Palmer, J.~Onton, R.~Oostenveld, and S.~Makeig, ``{Independent
  EEG sources are dipolar},'' \emph{PloS one}, vol.~7, no.~2, p. e30135, 2012.

\bibitem{oliva2001modeling}
A.~Oliva and A.~Torralba, ``Modeling the shape of the scene: A holistic
  representation of the spatial envelope,'' \emph{International journal of
  computer vision}, vol.~42, no.~3, pp. 145--175, 2001.

\bibitem{chevalier2004comparative}
P.~Chevalier, L.~Albera, P.~Comon, and A.~Ferr{\'e}ol, ``Comparative
  performance analysis of eight blind source separation methods on
  radiocommunications signals,'' in \emph{Proc. of IEEE International Joint
  Conference on Neural Networks}, vol.~1, 2004, pp. 273--278.

\bibitem{tichavsky2006performance}
P.~Tichavsky, Z.~Koldovsky, and E.~Oja, ``Performance analysis of the {FastICA}
  algorithm and cramer-rao bounds for linear independent component analysis,''
  \emph{IEEE transactions on Signal Processing}, vol.~54, no.~4, pp.
  1189--1203, 2006.

\end{thebibliography}

\end{document}